\documentclass{article}
\usepackage{spconf,amsmath,graphicx}
\usepackage{xspace}
\usepackage{times}
\usepackage{epsfig}
\usepackage{amssymb}

\usepackage{subfigure}
\usepackage{multirow}
\usepackage{enumitem}
\usepackage{booktabs}
\usepackage{color}
\usepackage{colortbl}
\definecolor{dkgreen}{rgb}{0,0.6,0}
\definecolor{gray}{rgb}{0.5,0.5,0.5}
\definecolor{mauve}{rgb}{0.58,0,0.82}
\definecolor{LightCyan}{rgb}{0.88,1,1}

\usepackage[pagebackref=true,breaklinks=true,letterpaper=true,colorlinks,bookmarks=false]{hyperref}

\newcommand{\mU}{\textbf{U}}
\newcommand{\mV}{\textbf{V}}

\newcommand{\mM}{\boldsymbol{M}}

\newcommand{\stkout}[1]{{\ifmmode\text{\sout{\ensuremath{#1}}}\else\sout{#1}\fi}}

\makeatletter
\DeclareRobustCommand\onedot{\futurelet\@let@token\@onedot}
\def\@onedot{\ifx\@let@token.\else.\null\fi\xspace}

\def\eg{\emph{e.g}\onedot} 
\def\ie{\emph{i.e}\onedot} 
\def\cf{\emph{c.f}\onedot} 
  
\def\wrt{w.r.t\onedot} 

\makeatother

\newcommand{\comment}[1]{}

\title{Flow Dynamics Correction for Action Recognition}
\name{Lei Wang\thanks{* This paper has been accepted for IEEE ICASSP 2024.}\textsuperscript{$*, \dagger,\S$} \qquad Piotr Koniusz\textsuperscript{$\S,\dagger$}
}
\address{$^{\dagger}$Australian National University, $^\S$Data61/CSIRO}

\begin{document}



\maketitle
\begin{abstract}
Various research studies indicate that action recognition performance highly depends on the types of motions being extracted and how accurate the human actions are represented. In this paper, we investigate different optical flow, and features extracted from these optical flow that capturing both short-term and long-term motion dynamics. 
We perform power normalization on the magnitude component of optical flow for flow dynamics correction to boost subtle or dampen sudden motions. We show that existing action recognition models which rely on optical flow are able to get performance boosted with our corrected optical flow. To further improve performance, we integrate our corrected flow dynamics into popular models through a simple hallucination step by selecting only the best performing optical flow features, and we show that by `translating' the CNN feature maps into these optical flow features with different scales of motions leads to the new state-of-the-art performance on several benchmarks including HMDB-51, YUP++, fine-grained action recognition on MPII Cooking Activities, and large-scale Charades.
\end{abstract}


\section{Introduction}
\label{sec:intro}

The motion cues for Action Recognition (AR)~\cite{lei_icip_2019, piotr2019, qin2022fusing, wang2022uncertainty, wang2022temporal, wang20233mformer, koniusz2021high, lei2023} can be extracted from multiple resources, \eg, RGB videos, depth videos, 3D point clouds, skeleton sequences, and optical flow videos. A thorough comparison of using different kinds of features for AR can be found in review papers~\cite{lei_thesis_2017, lei_tip_2019, lei2023}. Many state-of-the-art (SOTA) AR methods, apart from the use of RGB frames, rely on some form of optical flow which comes in many flavours. The introduction of optical flow estimation~\cite{tvl1_opt, ldof, deepflow, epicflow} led to a dramatic boost of performance in many areas of AR. 
TV-L1~\cite{tvl1_opt} preserves the discontinuities in the flow field, and provides an increased robustness in terms of occlusions, illumination changes and noise. LDOF~\cite{ldof} integrates rich descriptors into the variational optical flow setting to cope with large displacements. 
DeepFlow~\cite{deepflow}  boosts the performance \wrt fast motions by employing LDOF with a descriptor matching within a multi-stage architecture. 
EpicFlow~\cite{epicflow} targets large displacements with significant occlusions through a dense matching by edge-preserving interpolation from a sparse set of matches. 
These optical flow computation methods are quite mature and widely used in practice, hence they are of our interest for further investigations.

\begin{figure}[t]
\subfigure[{\em Marathon}: $\text{stride}=$1, 4, 8 and 12 respectively (from left to right). ]{\label{fig:yup-static-stride}
\includegraphics[trim=1.5cm 1.5cm 1.5cm 1.5cm, clip=true, width=.245\linewidth]{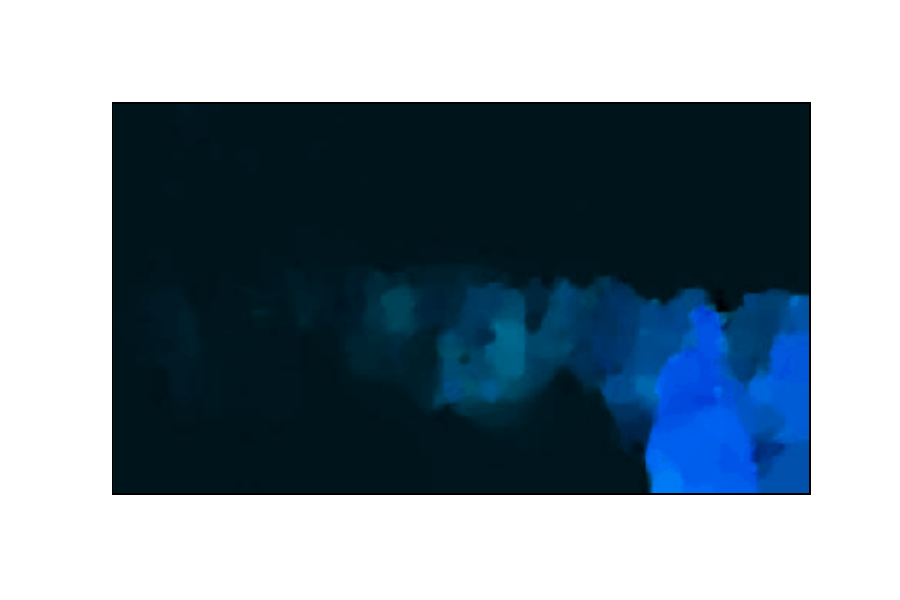}\hfill
\includegraphics[trim=1.5cm 1.5cm 1.5cm 1.5cm, clip=true,width=.245\linewidth]{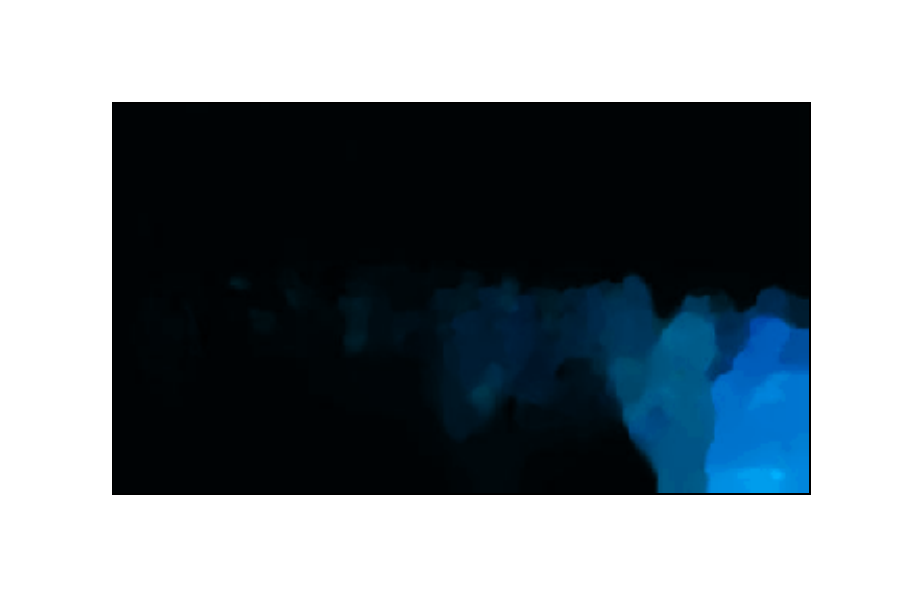}\hfill
\includegraphics[trim=1.5cm 1.5cm 1.5cm 1.5cm, clip=true,width=.245\linewidth]{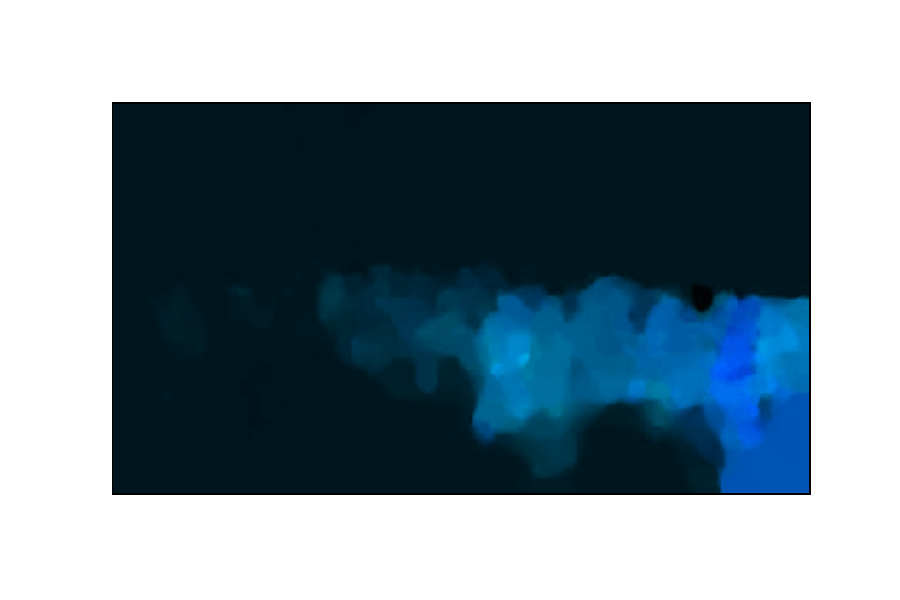}\hfill
\includegraphics[trim=1.5cm 1.5cm 1.5cm 1.5cm, clip=true,width=.245\linewidth]{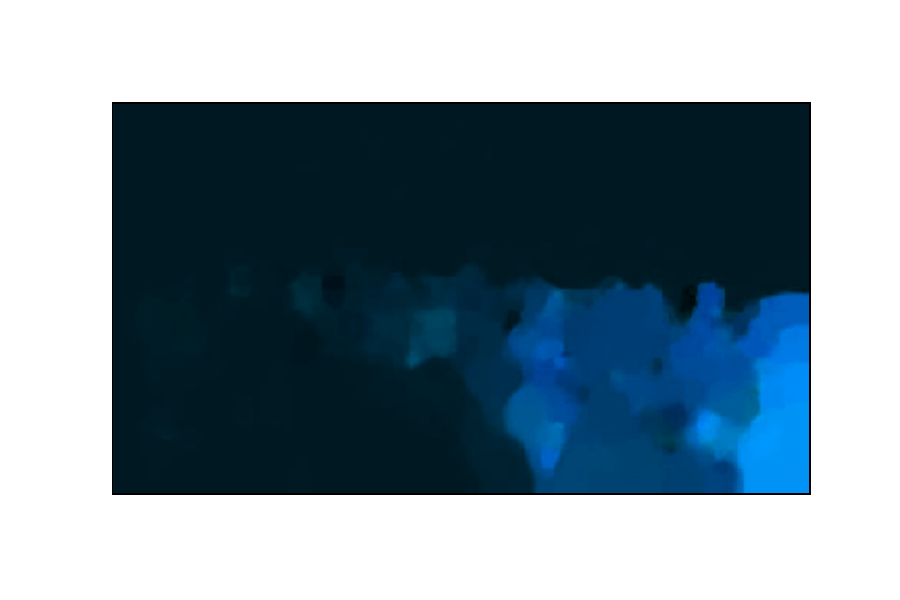}\\
}
\subfigure[{\em Kick ball}: $\text{stride}=$1, 2 and 4.]{\label{fig:hmdb51-stride-1}
\includegraphics[trim=3.0cm 1.5cm 3.0cm 1.5cm, clip=true,width=.162\linewidth]{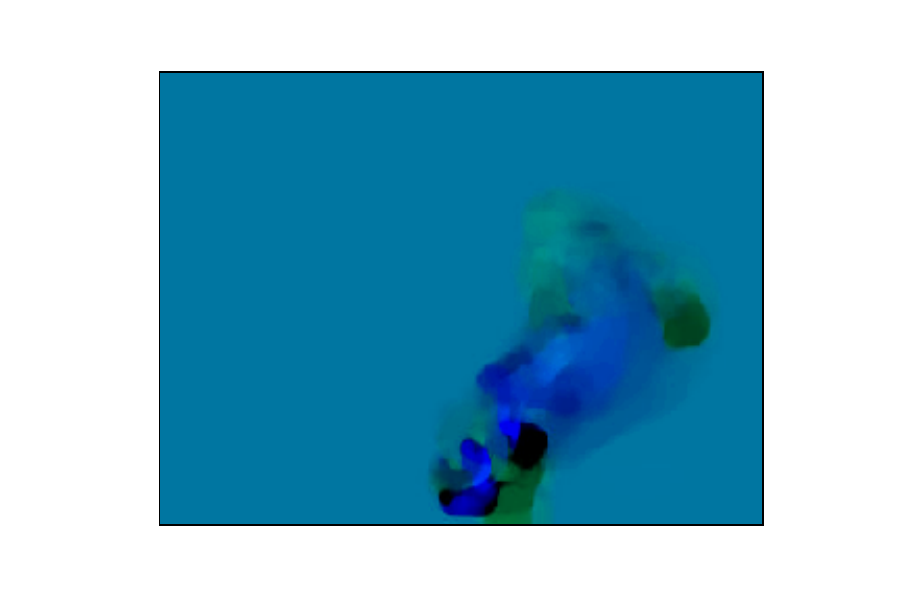}\hfill
\includegraphics[trim=3.0cm 1.5cm 3.0cm 1.5cm, clip=true,width=.162\linewidth]{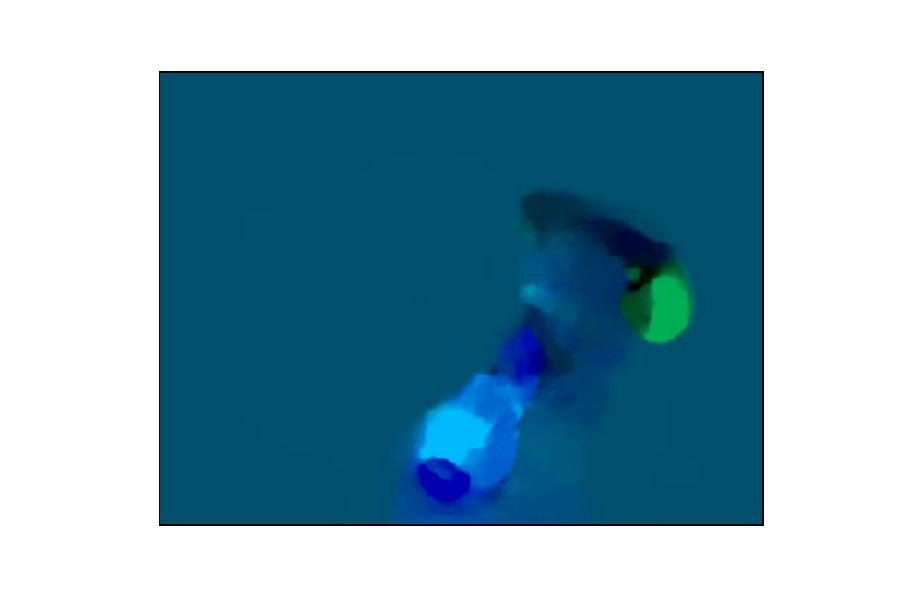}\hfill
\includegraphics[trim=3.0cm 1.5cm 3.0cm 1.5cm, clip=true,width=.162\linewidth]{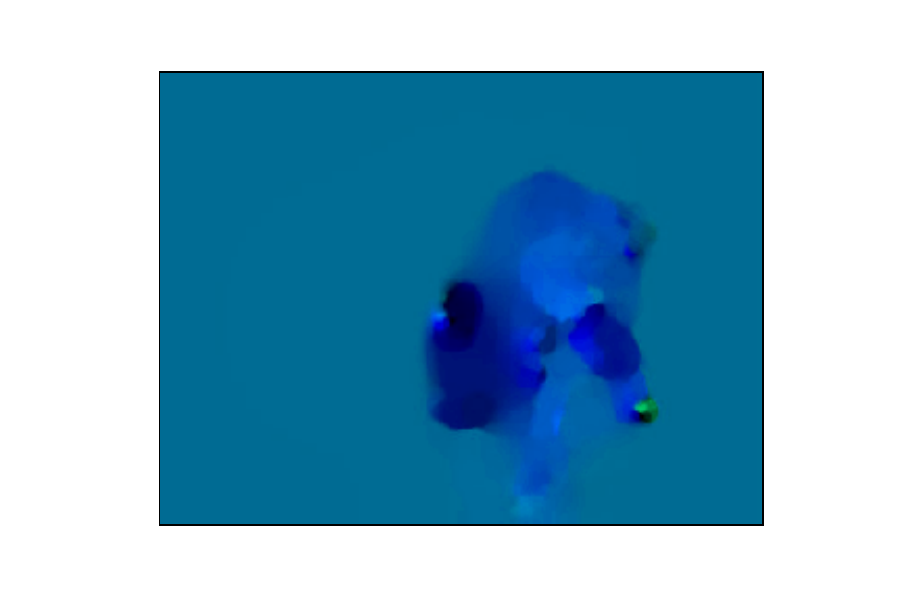}
}
\subfigure[{\em Situp}: $\text{stride}=$1, 2 and 4.]{\label{fig:hmdb51-stride-2}
\includegraphics[trim=3.0cm 1.5cm 3.0cm 1.5cm, clip=true,width=.162\linewidth]{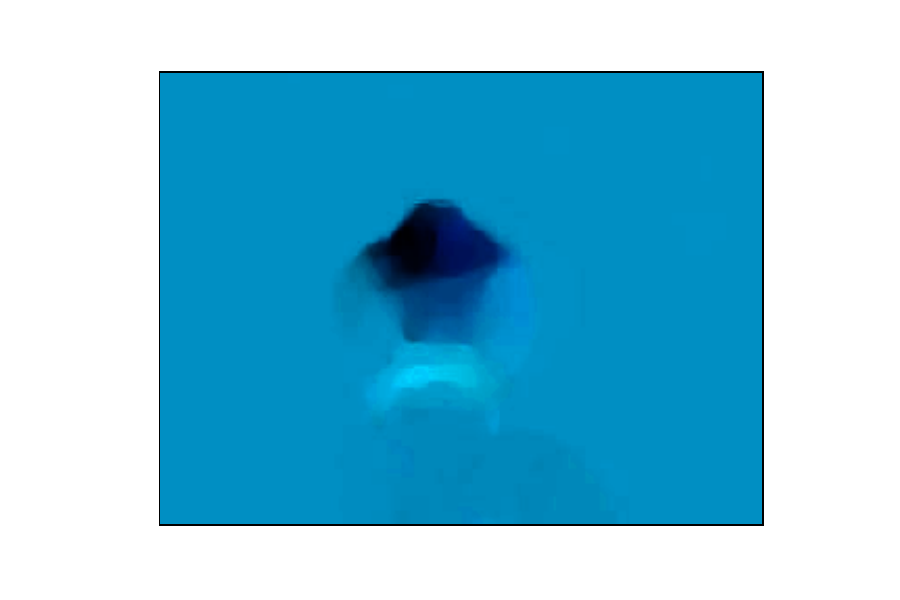}\hfill
\includegraphics[trim=3.0cm 1.5cm 3.0cm 1.5cm, clip=true,width=.162\linewidth]{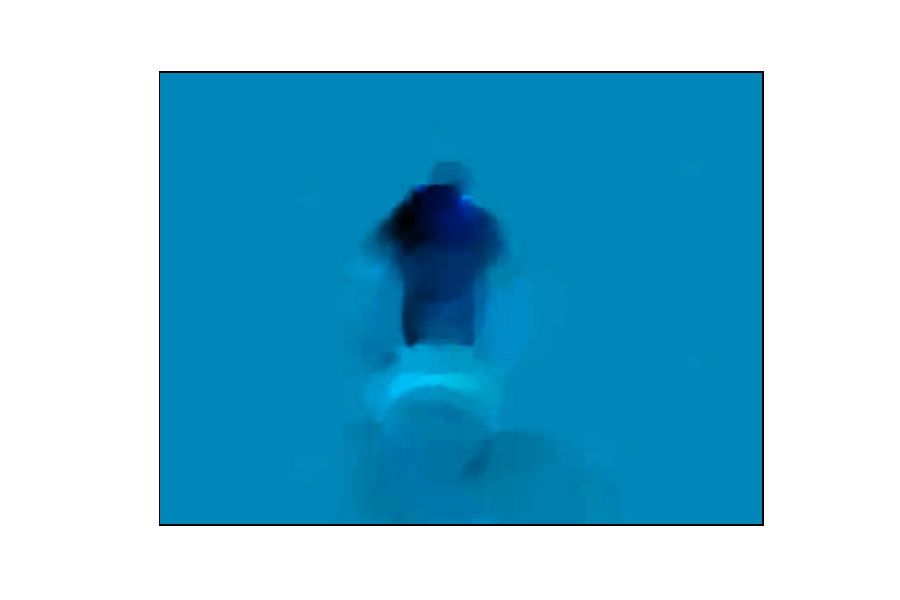}\hfill
\includegraphics[trim=3.0cm 1.5cm 3.0cm 1.5cm, clip=true,width=.162\linewidth]{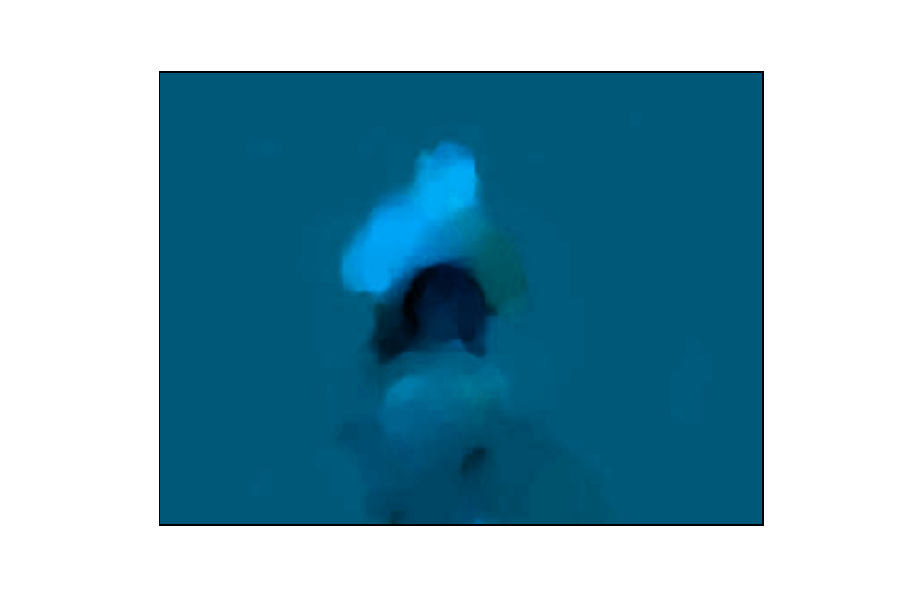}
}
\vspace{-0.5cm}
\caption{Multi-stride optical flow (LDOF) on (a) YUP++ and (b) -- (c) HMDB-51. Different strides (temporal scales) can capture different granularity levels of motions, and the visual appearance varies between different temporal scales.}
\label{fig:dynamics}
\vspace{-0.5cm}
\end{figure}

\begin{figure}[t]
\subfigure[DeepFlow: $\gamma=$0.1, 0.5 and 5.]{\label{fig:hmdb51-fc1}
\includegraphics[trim=3.0cm 1.5cm 3.0cm 1.5cm, clip=true,width=.162\linewidth]{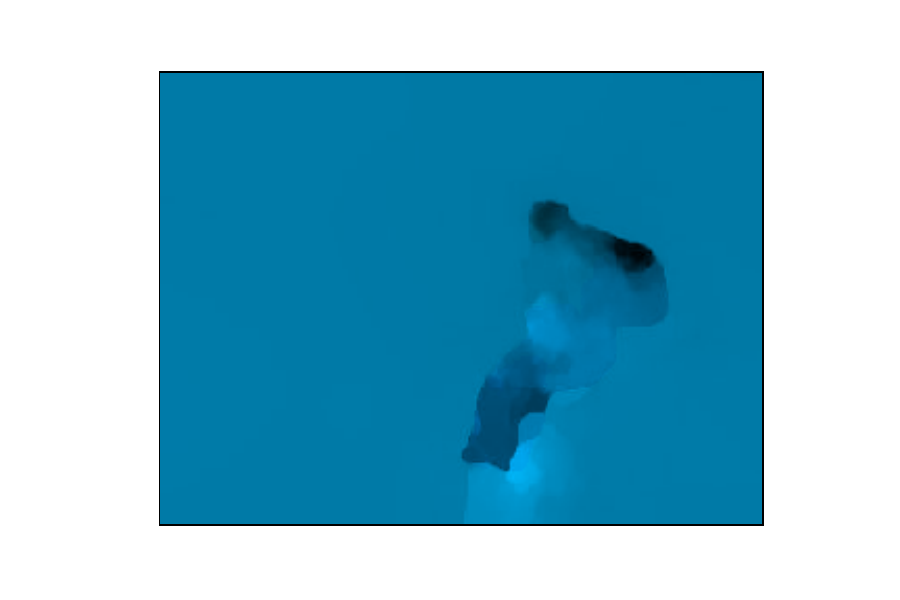}\hfill
\includegraphics[trim=3.0cm 1.5cm 3.0cm 1.5cm, clip=true,width=.162\linewidth]{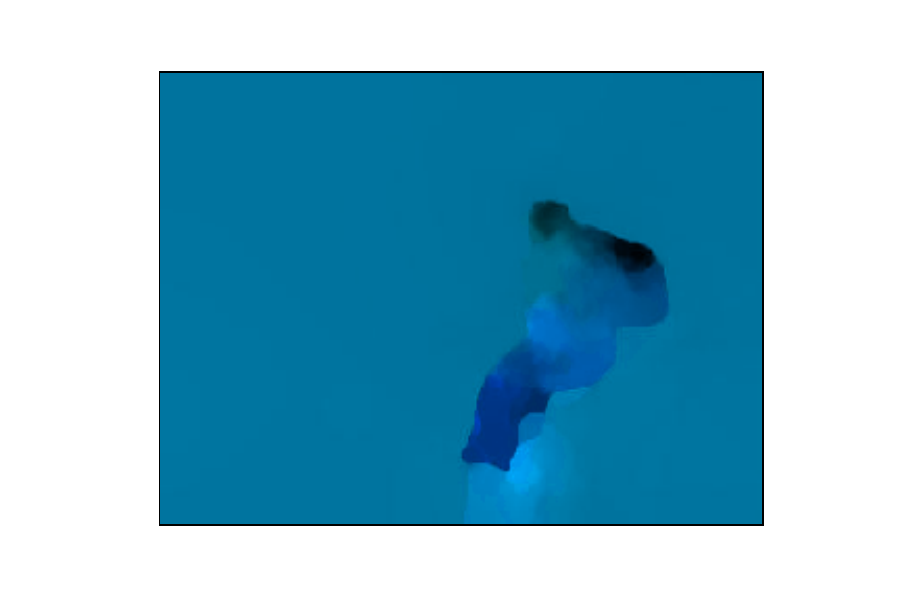}\hfill
\includegraphics[trim=3.0cm 1.5cm 3.0cm 1.5cm, clip=true,width=.162\linewidth]{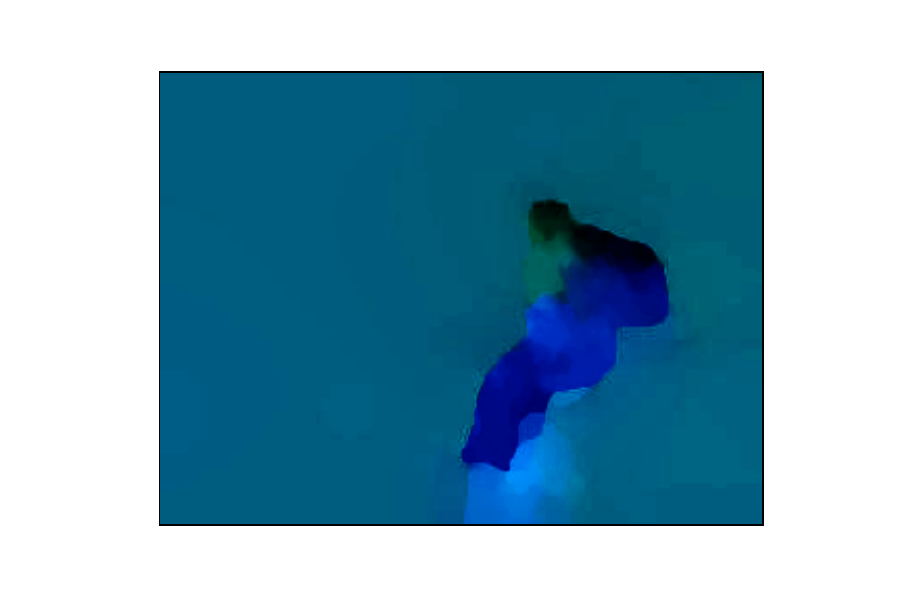}
}
\subfigure[TV-L1: $\gamma=$0.1, 0.5 and 5.]{\label{fig:hmdb51-fc2}
\includegraphics[trim=3.0cm 1.5cm 3.0cm 1.5cm, clip=true,width=.162\linewidth]{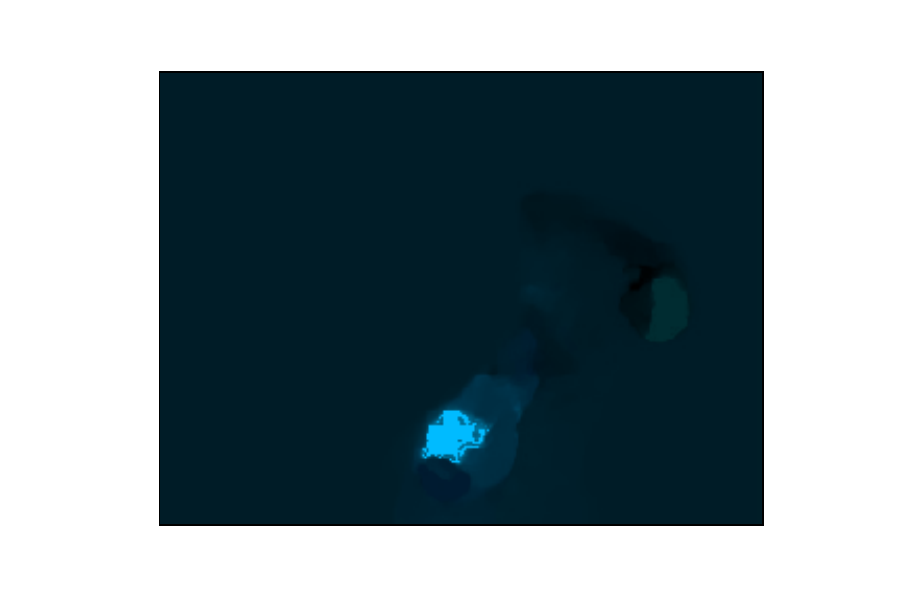}\hfill
\includegraphics[trim=3.0cm 1.5cm 3.0cm 1.5cm, clip=true,width=.162\linewidth]{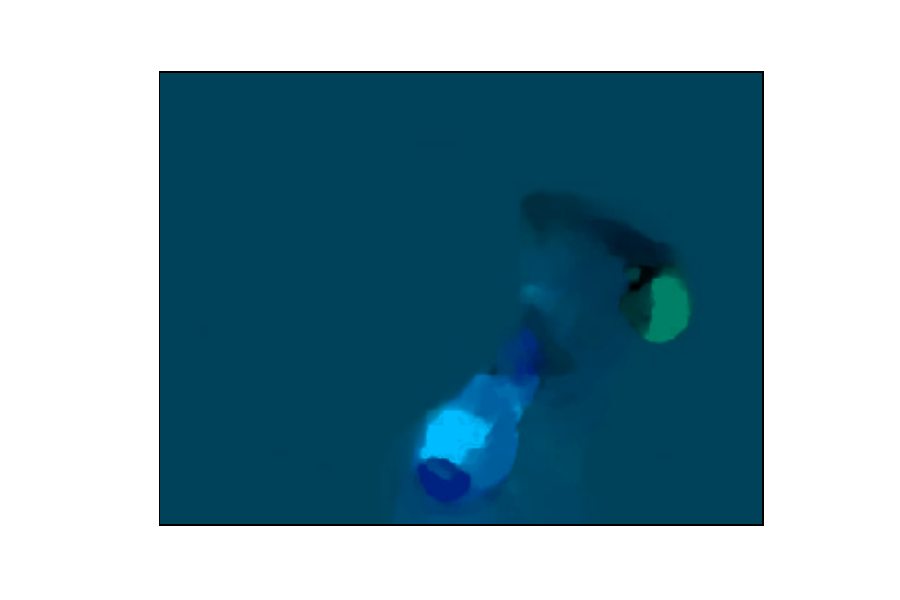}\hfill
\includegraphics[trim=3.0cm 1.5cm 3.0cm 1.5cm, clip=true,width=.162\linewidth]{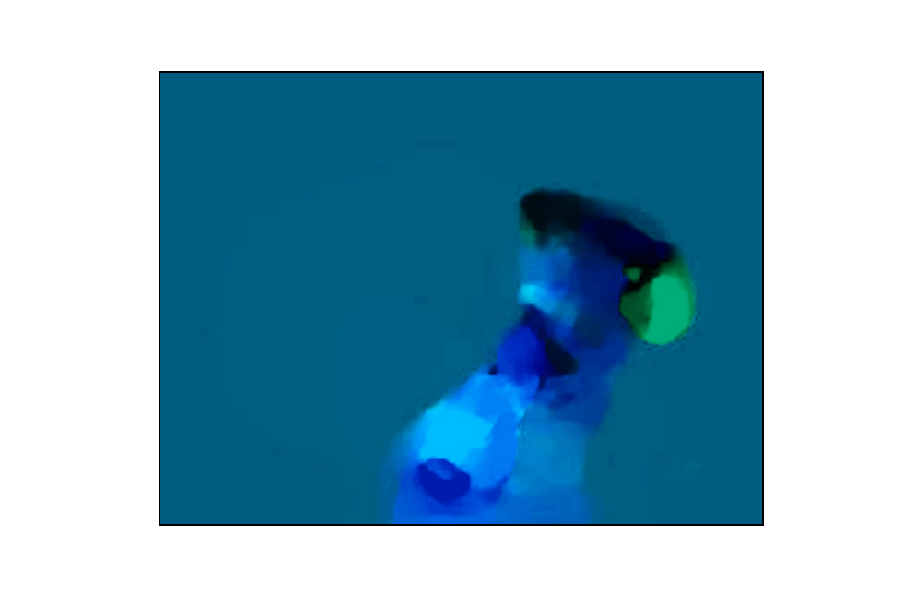}
}
\subfigure[EpicFlow: $\gamma=0.1$.]{\label{fig:hmdb51-fc3}
\includegraphics[trim=3.0cm 1.5cm 3.0cm 1.5cm, clip=true,width=.16\linewidth]{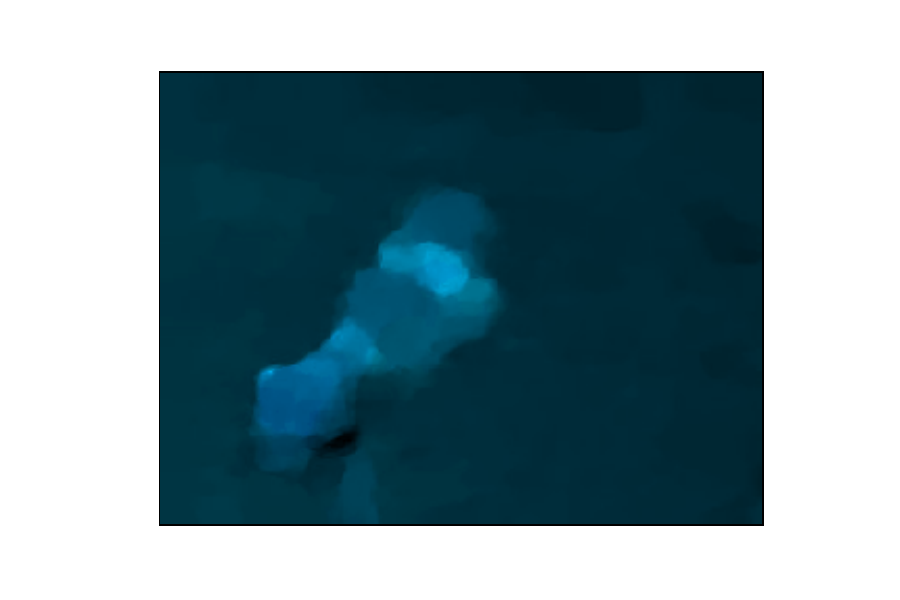}\hfill
\includegraphics[trim=3.0cm 1.5cm 3.0cm 1.5cm, clip=true,width=.16\linewidth]{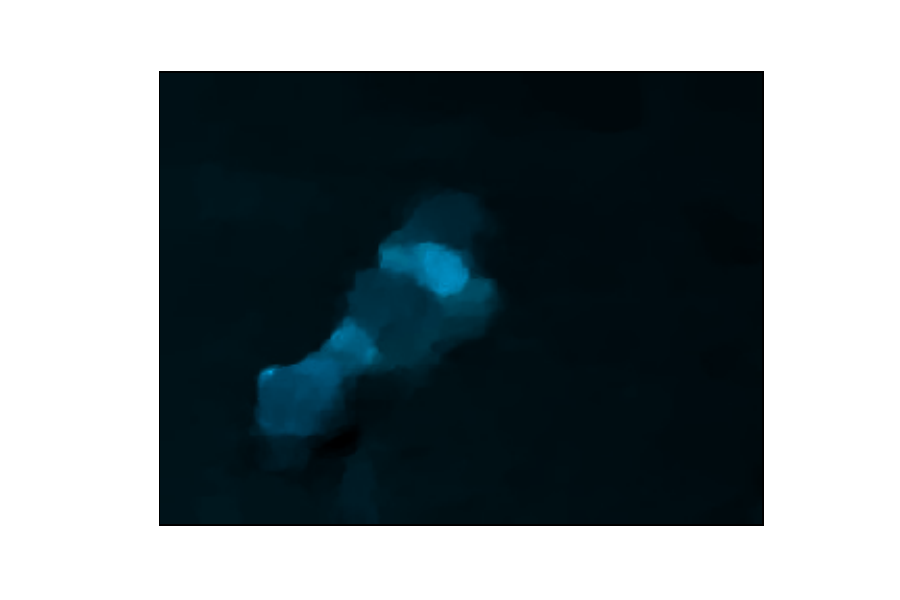}\hfill
}
\subfigure[EpicFlow: $\gamma=0.5$.]{\label{fig:hmdb51-fc4}
\includegraphics[trim=3.0cm 1.5cm 3.0cm 1.5cm, clip=true,width=.16\linewidth]{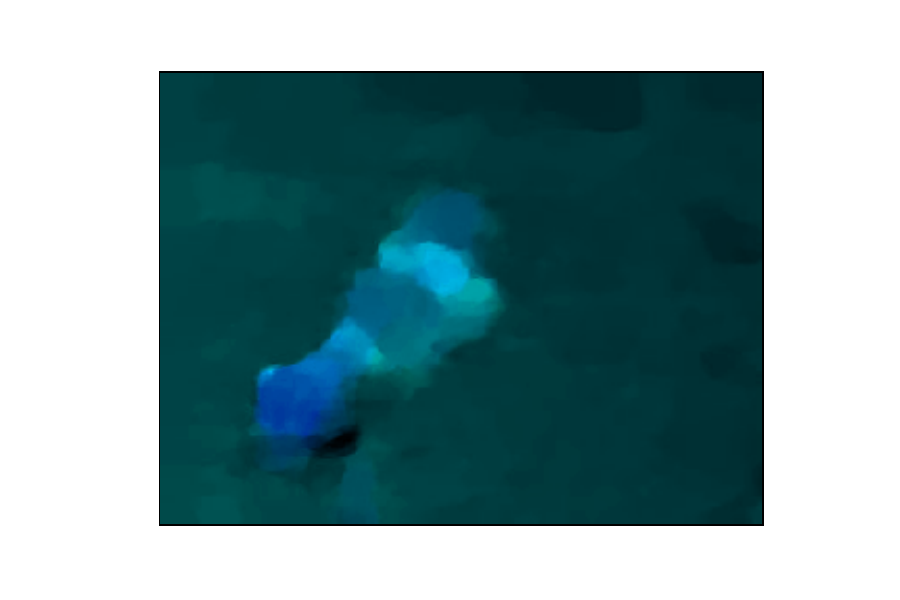}\hfill
\includegraphics[trim=3.0cm 1.5cm 3.0cm 1.5cm, clip=true,width=.16\linewidth]{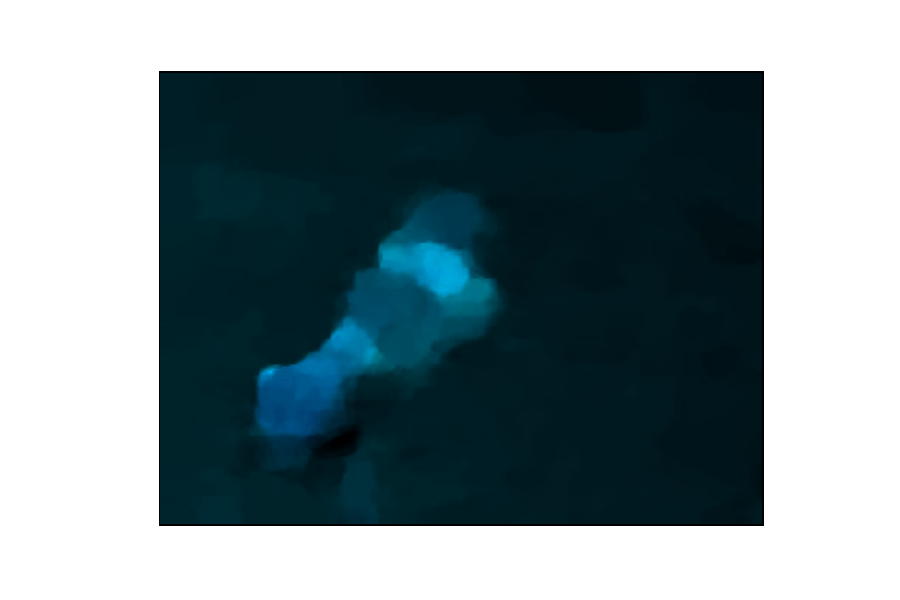}\hfill
}
\subfigure[EpicFlow: $\gamma=5$.]{\label{fig:hmdb51-fc5}
\includegraphics[trim=3.0cm 1.5cm 3.0cm 1.5cm, clip=true,width=.16\linewidth]{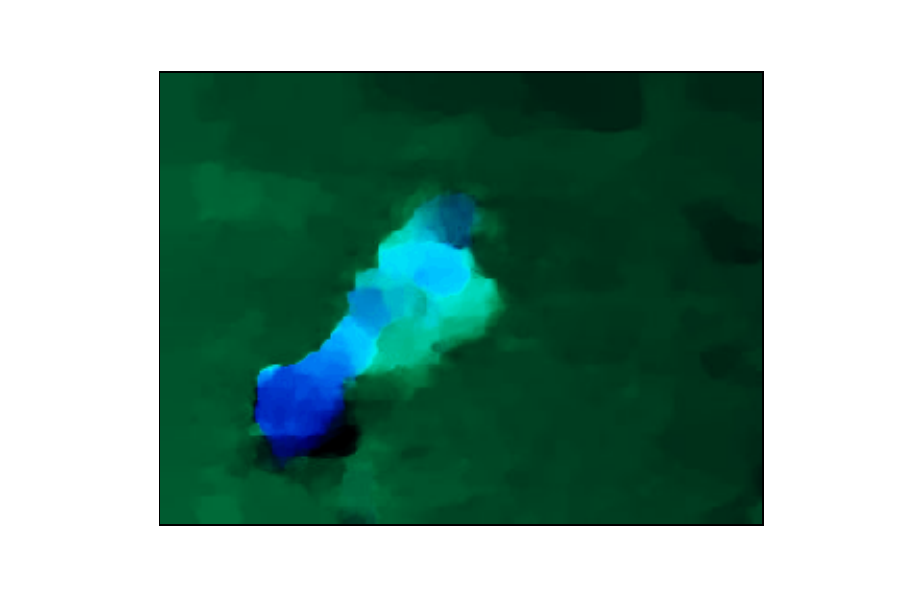}\hfill
\includegraphics[trim=3.0cm 1.5cm 3.0cm 1.5cm, clip=true,width=.16\linewidth]{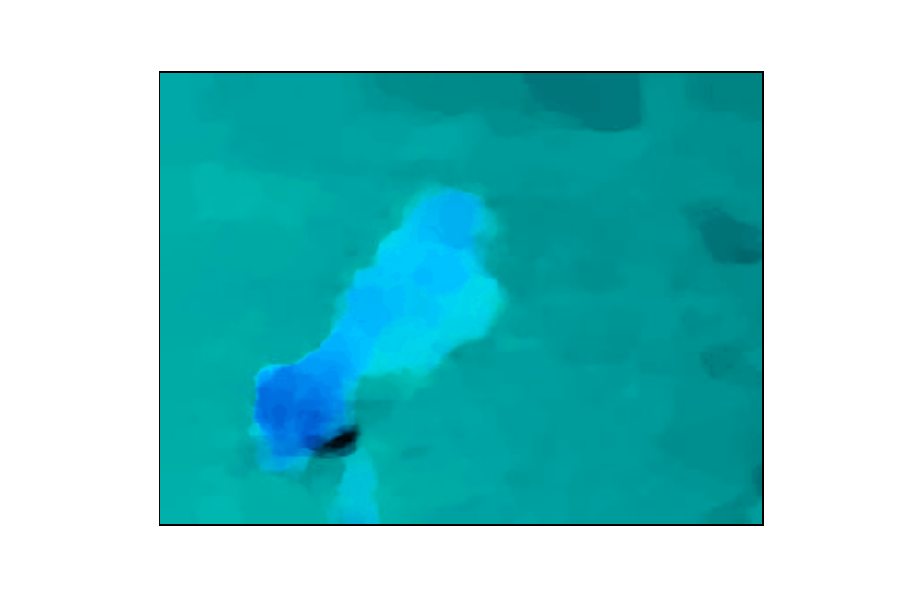}
}
\vspace{-0.5cm}
\caption{
(a)--(b) show the strength of PN ($\gamma$) for optical flow correction on action {\em Kick ball}. Small $\gamma$ preserves the dominant motions and large $\gamma$ boosts some weak motions and maintains more rich motion dynamics. \!Each pair of figures in (c) -- (e) 
shows with (left) and without (right) dominant motions on action {\em dribble}. 
\!All actions are from HMDB-51.}
\label{fig:flow-correction}
\vspace{-0.5cm}
\end{figure}

However, in general, articulated motion and human motion in particular are problematic. Some human body parts, \ie, hands can move very fast, whereas other parts may follow a slower motion pattern. Indeed, different motion speeds likely introduce nuisance variations that contribute to poorer recognition results.
We perform power normalization on the magnitude component of optical flow for the correction of flow dynamics to boost subtle or dampen sudden motions.
We show that with our corrected flow dynamics, handcrafted IDT descriptors~\cite{improved_traj}, popular two-stream network~\cite{christoph2016cvpr}, I3D~\cite{i3d_net} and even AssembleNet/AssembleNet++~\cite{assemblenet,assemblenet_plus} are able to improve the AR performance by 3 -- 5\% on average.
Since these SOTA AR approaches rely on optical flow estimation methods to pre-compute motion information for CNNs and such a two-stage method is computationally expensive, storage demanding, and not end-to-end trainable,
%
we also propose simple trainable CNN streams on top of a CNN network (\eg, I3D~\cite{i3d_net} and AssembleNet++~\cite{assemblenet_plus}) that learn to `translate' the RGB output into several OFFs that are extracted at different scales to form the short-term and long-term motion dynamics. Different OFFs are not only synthesized but they also provide self-supervisory signals.    
Our main contributions are:
\renewcommand{\labelenumi}{\roman{enumi}.}
\vspace{-0.1cm}
\hspace{-1.0cm}
\begin{enumerate}[leftmargin=0.6cm]
    \item We show that correcting optical flow maps by so-called power normalization (PN) produces various motion dynamics that dampen sudden motions or noise and magnify tiny motions. 
    \vspace{-0.2cm} 
    \item We investigate various aspects of our model, \eg, different kinds of optical flow or scales of motion (short-term and long-term motions). With the corrected flow dynamics, our model outperforms previous approaches on 4 benchmarks including dynamic scenes classification and fine-grained AR by a large margin. 
    \vspace{-0.2cm}
    \item We introduce a Selector for selecting the best corrected motion dynamics to learn the feature streams. We also show that different optical flow features (OFFs) extracted from either short-term or long-term motion dynamics can be synthesized implicitly to handle various speeds and dynamics of actions.
\end{enumerate}

\section{Approach}
\label{sec:approach}



\vspace{-0.2cm}

\subsection{Flow Dynamics Correction}




\noindent{\bf Multi-stride optical flow computation.} We choose  TV-L1~\cite{tvl1_opt}, LDOF~\cite{ldof}, DeepFlow~\cite{deepflow} and EpicFlow~\cite{epicflow} because (i) they are often used in video classification tasks and (ii) they cope with large displacements, occlusions, and small motions. 
Setting $\text{stride} = 1$ is the most common setting that is widely used in the optical flow computations to capture the temporal information. 
In this work, we explore the effects of using different strides for optical flow computations. 

We let the stride step take values between one and the average number of frames in each dataset to form different scales of motion dynamics. On  HMDB-51 and YUP++, we use $\text{stride} = 1, 2, 4, 6, 8, 12, 15, 30, 45$ for all 4 optical flow computations. If the selected stride value is greater than the total number of frames in a given video, we drop this stream as the temporal information can be captured later in the shorter-term streams with smaller strides. 
Fig.~\ref{fig:dynamics} shows some visualizations of LDOF with different strides to form different motion dynamics on YUP++ and HMDB-51. As shown in these figures, the multi-stride flow dynamics are very different from the original optical flow computed by the use of common setting ($\text{stride} = 1$). Optical flow computed at different temporal scales are able to capture motion dynamics at different granularity levels, and the visual appearances in these optical flow are different; hence, our multi-stride optical flow can capture more rich motion and related appearance information for downstream video processing tasks.

\begin{figure}[t]
\centering\includegraphics[trim=4.0cm 7.8cm 3.5cm 7.2cm, clip=true,width=\linewidth]{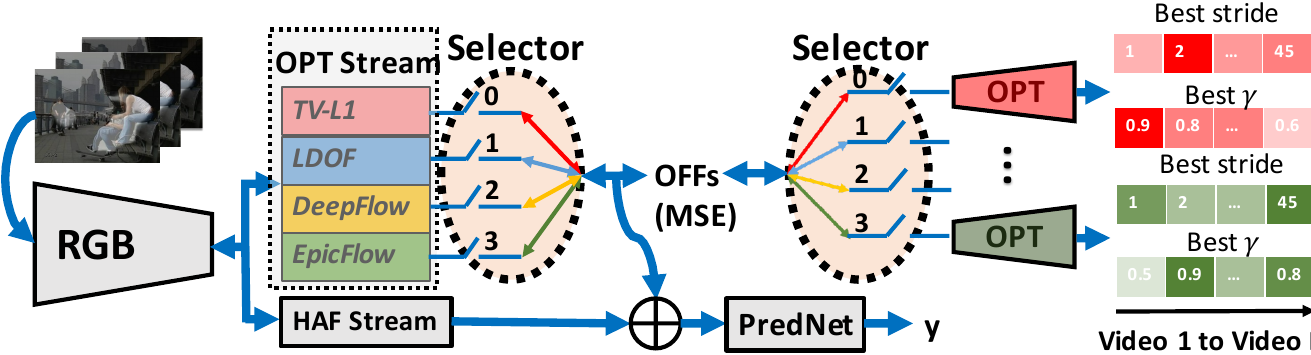}
\vspace{-0.8cm}
\caption{
We use optical flow (OPT) streams and a Selector to learn to hallucinate the best optical flow features (OFFs). The OFFs and features from the High Abstraction Features (HAF) stream are concatenated by $\bigoplus$, and then feed into the PredNet (a simple MLP) for classification. The Selector is used to choose the optical flow types we learn to hallucinate, given 
the best OFFs. MSE represents the mean square error loss, while $y$ denotes the output class label from PredNet.
}
\vspace{-0.2cm}
\label{fig:opt_hal}
\end{figure}

\begin{figure}[t]
\centering\includegraphics[trim=2.0cm 7.3cm 2.5cm 6.2cm, clip=true,width=\linewidth]{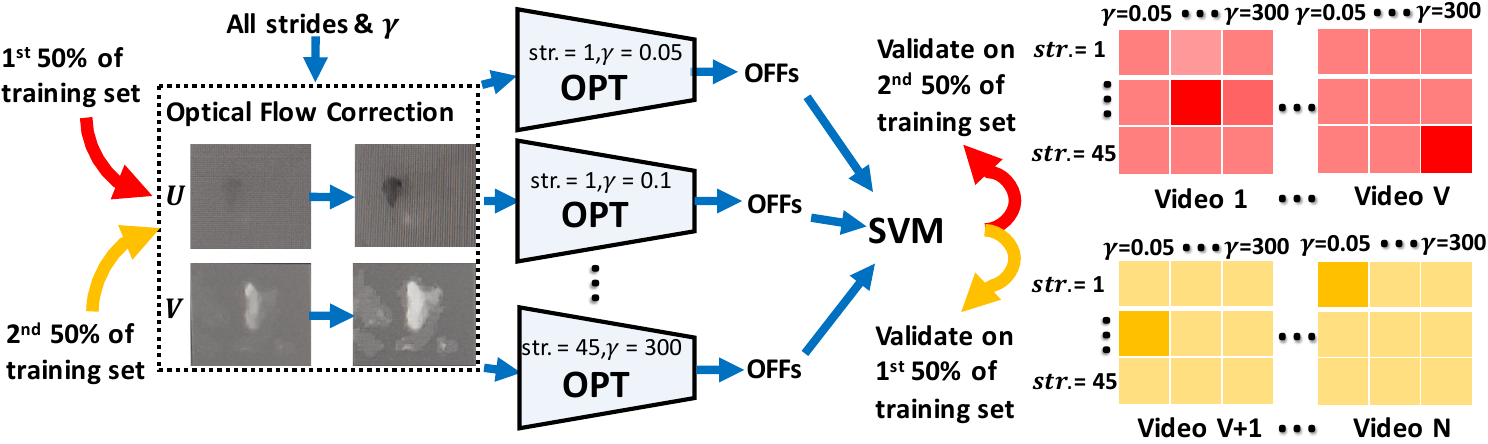}
\vspace{-0.8cm}
\caption{Our stride and $\gamma$ selector. Pre-selection of best (stride, $\gamma$) per optical flow type per video. 
}
\vspace{-0.3cm}
\label{fig:opt_hal2}
\end{figure}

\noindent{\bf Optical flow correction.} Let $\mU$ and $\mV$ be two maps with the displacement components (along $x$ and $y$ axis, respectively) of the computed multi-stride optical flow. The magnitude and angle of the optical flow $(\mU, \mV)$ are computed (by element-wise operations) as
\vspace{-0.2cm}
{
\begin{align}
\mM = \sqrt{\mU^2 + \mV^2},\\
\boldsymbol{\Phi} = \text{arctan}(\mU/\mV).
\label{eq:mag_ang}
\end{align}
}
\noindent As videos are highly affected by many issues like noise, camera shaking, dynamic background environments and a mixture of fast and slow motions, \eg, human actions, we apply the element-wise power normalization (PN), on the magnitude component $\mM$ for the flow correction to get the power normalized magnitude matrix $\mM'$
\vspace{-0.2cm}
{
\begin{align}
 \mM' = \text{sign}(\mM) \cdot (1-(1-\text{abs}(\mM))^\gamma), 
\label{eq:mag_corr}
\end{align}
}
where $\gamma>0$ decides the strength of PN, and all operations are element-wise. 
 The PN here is used for the flow correction that is performed on each optical flow frame. The normalization is done on the magnitude component of the optical flow so as to boost or dampen subtle or sudden motions. We then compute optical flow features (OFFs) from such mended motion clips. 
 The use of abs and sign in Eq.~\eqref{eq:mag_corr} is for maintaining the motion direction. 
 We use $\gamma>1$ to boost weak and dampen dominant motions (\cf $0<\gamma<1$ to preserve only dominant motions) which gives us selective focus on various motion dynamics. Note that if $\gamma = 1$, PN is not performed. 
 Finally, we recover two optical flow maps $(\mU', \mV')$ based on the corrected $\mM'$ and $\boldsymbol{\Phi}$ as 
 \vspace{-0.2cm}
 {
\begin{align}
 \mU' = \mM' \cdot \text{sin}(\boldsymbol{\Phi}),\\
 \mV' = \mM' \cdot\text{cos}(\boldsymbol{\Phi}). 
\label{eq:new_flow}
\end{align}
}

\vspace{-0.2cm}
Fig.~\ref{fig:flow-correction} shows some visualizations of corrected flow dynamics on HMDB-51. The color intensity shows the effects of power normalization with different $\gamma$ values. We notice that smaller $\gamma$ preserves mainly the dominant motions whereas large $\gamma$ boosts some weak motions and keeps more rich and fine-grained information.
\vspace{-0.2cm}
\subsection{Stride and $\gamma$ selector} 

We introduce a lightweight hallucination\footnote{`Hallucination' conveys the model's ability to generate video representations during the test stage, making them available without the original, time-consuming computation and processing steps. } model (HAL) inspired by~\cite{Wang_2019_ICCV, leimm21}; however, our HAL only has the optical flow streams built on top of a backbone network. The input to our HAL is the RGB video, and it learns (during training) to translate latent features from RGB into OFFs, which represent various motion dynamics based on optical flow. 
Our pipeline uses the corrected flow dynamics illustrated in Fig.~\ref{fig:opt_hal}. 
There are 4 switches that activate the corresponding optical flow streams based on the selection made by the Selector. 


Figure \ref{fig:opt_hal2} shows our stride and $\gamma$ selector.
Given the corrected optical flow, we split train data into two halves. We train scoring  optical flow networks (\eg, I3D or AssembleNet/AssembleNet++ optical flow stream pre-trained on Kinetics-400), one per optical flow type, stride choice and $\gamma$ choice. We train on one half of train data, and score via SVM each video on the second half of train data in terms of which (stride, $\gamma$) recognises video correctly (or is the closest to correct decision). Then, we train networks on the second half of the train data and score videos on the first half. With such scoring, we can train four optical flow networks by directing to them best (stride, $\gamma$) per video. 
We choose  
(i) the best performing optical flow feature for optimal (stride, $\gamma$) per optical flow type or 
(ii) only one best performing optical flow type to hallucinate thus  preventing the overparametrization (Selector uses pre-scores to choose also the best optical flow type, so we pre-select (type, stride, $\gamma$) per video). 
As a result, our proposed method is able to generate the better OFFs without the need of optical flow computation during the test stage. Due to the optical flow type and best (stride, $\gamma$) selectors, the network generates features with the best motion dynamics per video rather than static features from one kind of optical flow and fixed stride. 

\section{Experiments}
\label{sec:exper}
\vspace{-0.2cm}
\subsection{Datasets and Protocols}
\label{sec:data}

We evaluate the use of flow dynamics correction in popular action recognition models on 4 benchmarks: HMDB-51~\cite{kuehne2011hmdb}, YUP++~\cite{yuppp}, MPII Cooking Activities~\cite{rohrbach2012database} and Charades~\cite{sigurdsson2016hollywood}. 
Using standard protocols, we report recognition/classification accuracy (\%) for HMDB-51 and YUP++, mean average precision (mAP) for MPII and Charades.
First, we use our HAL with flow dynamics correction for ablation studies, and then we compare our method versus the SOTA methods.

\begin{figure}[t]
\centering
\vspace{-0.1cm}
\begin{tabular}[t]{cc}

\subfigure[]
{\label{fig:opts_comp_hmdb}\includegraphics[trim=0cm 0cm 0cm 0.5cm, clip=true, width=0.48\linewidth]{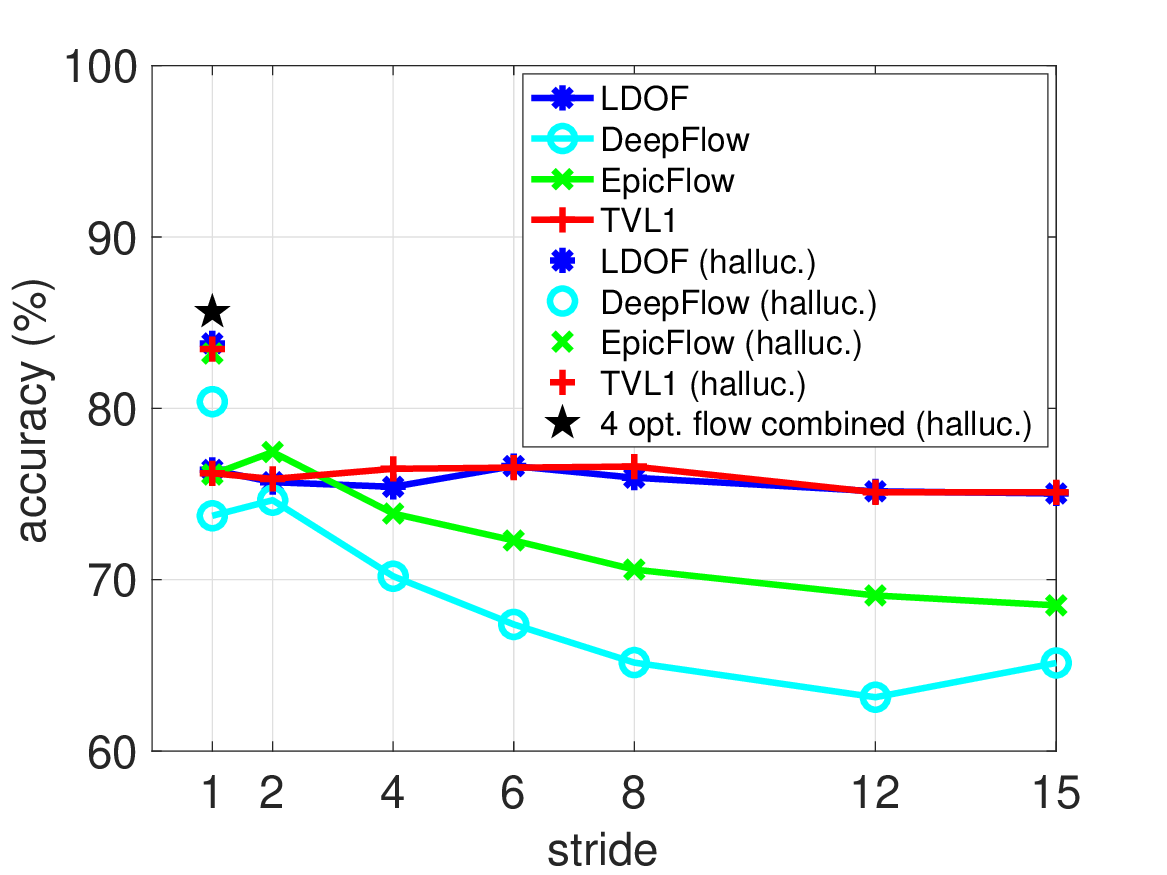}}
\subfigure[]
{\label{fig:opts_comp_yup}\includegraphics[trim=0cm 0cm 0cm 0.5cm, clip=true, width=0.48\linewidth]{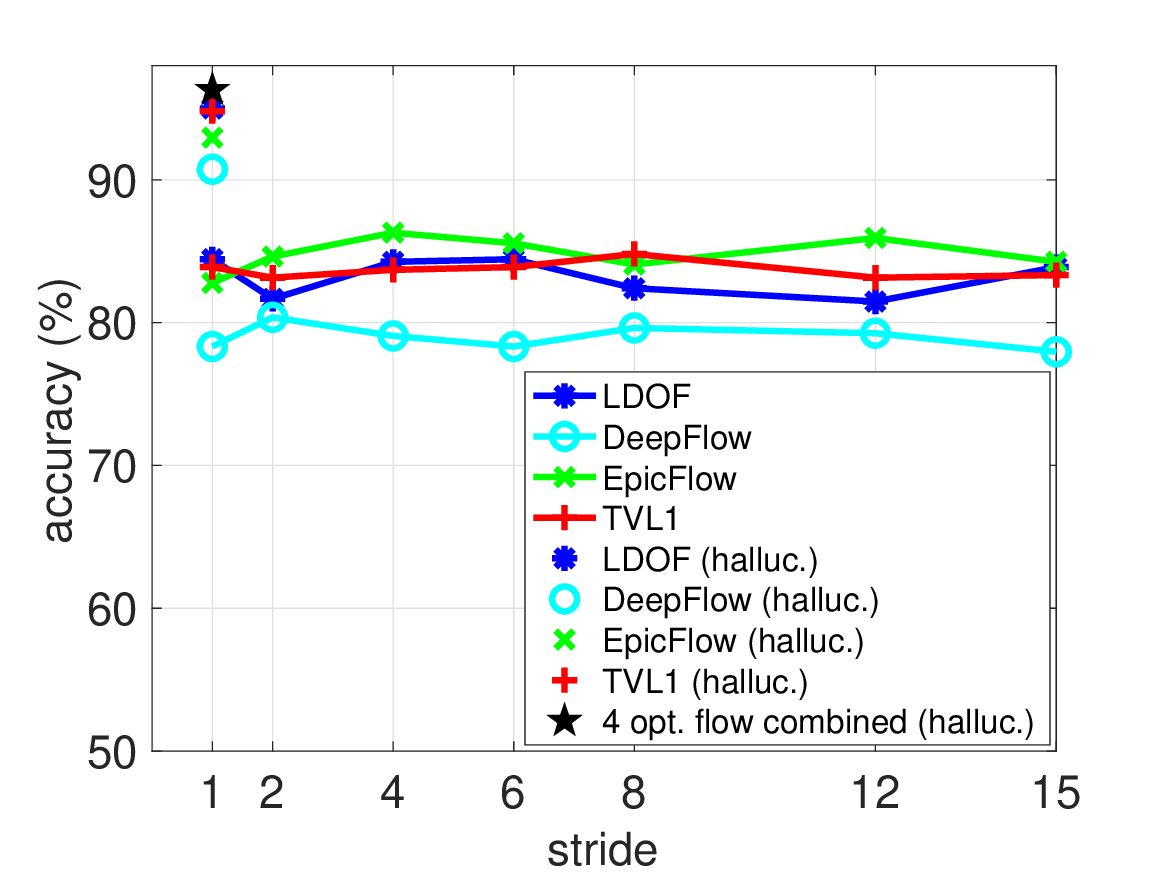}}\vspace{-0.35cm}\\
\subfigure[]
{\label{fig:dom_hmdb}\includegraphics[trim=0cm 0cm 0cm 0.5cm, clip=true, width=0.48\linewidth]{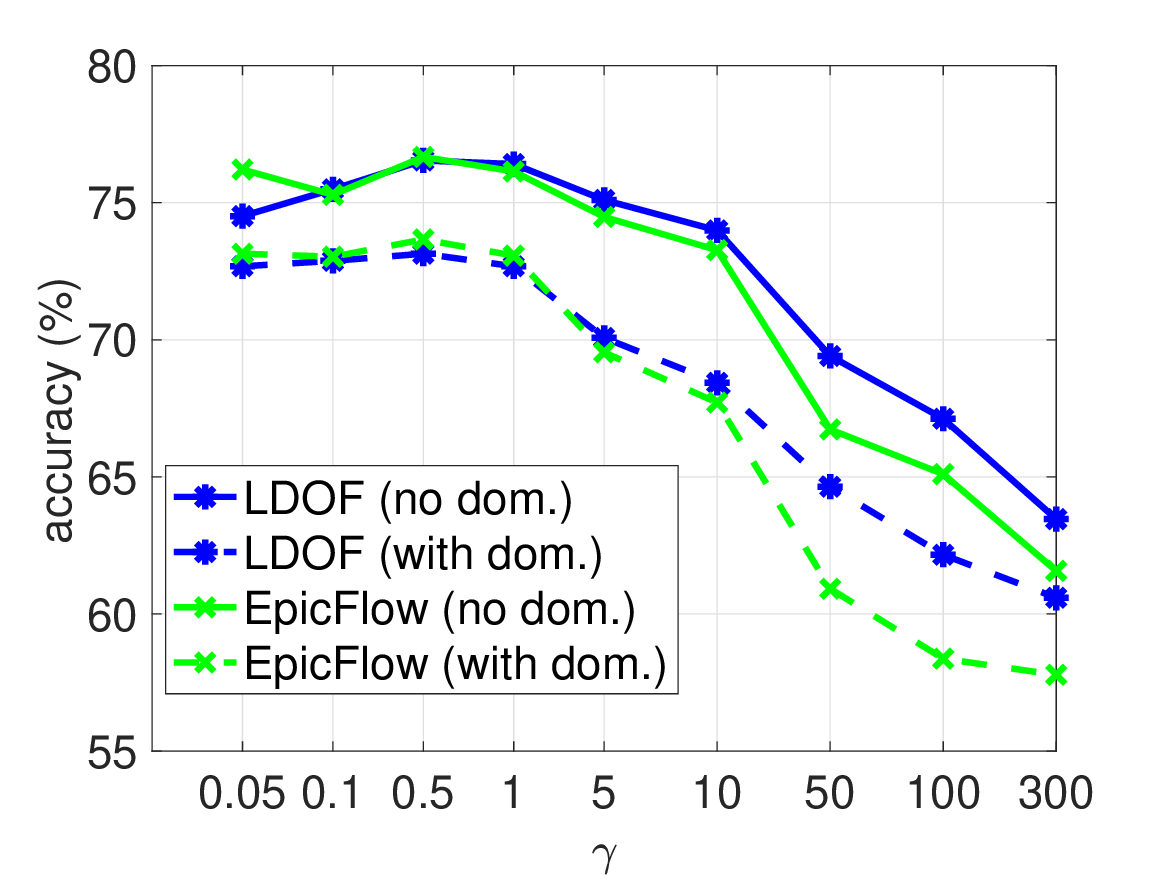}}
\subfigure[]
{\label{fig:dom_yup}\includegraphics[trim=0cm 0cm 0cm 0.5cm, clip=true, width=0.48\linewidth]{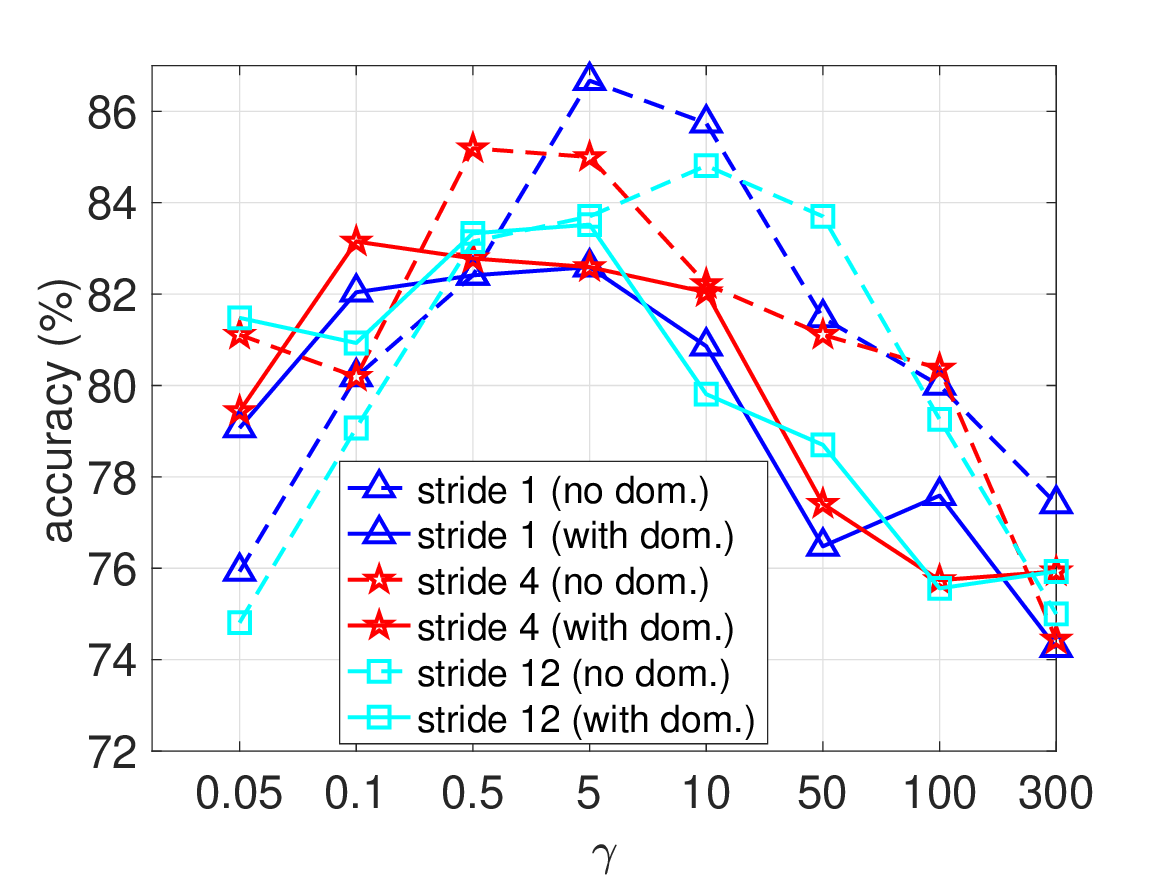}}

\end{tabular}
\vspace{-0.6cm}
\caption{(Top row) Evaluations of OFFs w.r.t. different scales of motions on ({\em a}) HMDB-51 (split 1) and ({\em b}) YUP++ ({\em static camera}). (Bottom row) Evaluations of PN with/without the use of dominant motions 
w.r.t. ({\em c}) different optical flow on HMDB-51 and ({\em d}) different scales of motions on YUP++.}
\vspace{-0.5cm}
\label{fig:opts_comp}
\end{figure}

\vspace{-0.2cm}

\subsection{Ablation Study}
\label{sec:ablation-study}

%

\noindent{\textbf{Flow estimation quality.}} 
Fig.~\ref{fig:opts_comp} (top) shows the comparison of using different OFFs on HMDB-51 and YUP++. As shown in Fig.~\ref{fig:opts_comp_hmdb}, the accuracies of using OFFs extracted from DeepFlow and EpicFlow decrease when long-term motion is used, whereas the TV-L1 and LDOF perform better in terms of both short-term and long-term motions. For natural dynamic scenes classification (see Fig.~\ref{fig:opts_comp_yup}), all OFFs perform almost equally well. This is mainly because the motions in natural dynamic scenes are generally periodic, whereas human actions are far more complicated in terms of the dynamics in different body parts. After integrating each optical flow feature into our hallucination pipeline, the performance increases by $\sim$ 9\% on HMDB-51 and $\sim$ 10\% on YUP++, respectively. Hallucinating all 4 OFFs further improves the performance by 2 -- 3\% on both datasets.

\noindent{\textbf{With dominant motions.}} Fig.~\ref{fig:opts_comp} (bottom) shows the performance comparison between with and without the use of dominant motions, \ie, dominant magnitude of $\mM$ is subtracted from it before PN. For AR on HMDB51 (see Fig.~\ref{fig:dom_hmdb}), using the dominant motions (with dom.) does not improve the overall recognition accuracies, and the performance for the use of LDOF and EpicFlow drops by 2-5\%. 
HMDB-51 is a challenging dataset with videos captured by moving cameras with noise, especially in sports-related actions, making environment-dependent actions a significant factor. Relying solely on dominant motions is insufficient for reliable action classification.
For natural dynamic scenes classification on the YUP++ dataset (see Fig.~\ref{fig:dom_yup}), using dominant motions (with dom.) when $\gamma \leq 0.5$ performs slightly better, as it filters out static camera motion.



\begin{table}[t]
\begin{center}
\setlength{\tabcolsep}{0.12em}
\renewcommand{\arraystretch}{0.70}
\fontsize{9}{9}\selectfont
\resizebox{0.65\linewidth}{!}{\begin{tabular}{ l c c c c }
\toprule
 & {\em sp1} & {\em sp2} & {\em sp3} & mean acc. \\
\hline
all 4 opt. flow (stride=1) & $ 83.5 $ & $ 83.5 $ & $ 83.5 $ & $ 83.5 $ \\
all 4 opt. flow (best stride)	& $85.6$ & $85.2$ & $85.5$ & $85.4$\\
\rowcolor{LightCyan}
all 4 opt. flow (corrected) & $87.5$ & $86.7$ & $87.5$ & $\mathbf{87.3}$\\
\rowcolor{LightCyan}
best opt. flow only (corrected)& $86.9$ & $86.8$ & $86.8$ & $86.8$\\
\bottomrule
\end{tabular}}
\vspace{-0.3cm}
\caption{Evaluations of our HAL variants on HMDB-51. sp1, sp2, and sp3 denote three splits.}
\label{tab:w-selector}
\end{center}
\vspace{-0.5cm}
\end{table}

\noindent{\textbf{With Selector.}} Tab.~\ref{tab:w-selector} shows the evaluations of our HAL variants. We first choose $\text{stride}\!=\!1$ for all 4 optical flow types and hallucinate all 4 OFFs on the HMDB-51 dataset. Note that this is the default setting which is widely used where the optical flow computation is done on two consecutive frames (we set it as baseline for comparison). As in Tab.~\ref{tab:w-selector}, choosing the best stride per optical flow ({\em all 4 opt. flow (best stride)}) performs better than using the common setting ($\text{stride}\!=\!1$) for all 4 OFFs by $\sim$ 2\%. Hallucinating the best stride and $\gamma$ per optical flow ({\em all 4 opt. flow (corrected)}) performs better than ({\em all 4 opt. flow (best stride)}) by $\sim$ 1.8\%. We also hallucinate the top performing optical flow feature choosing from all 4 OFFs by using our stride and $\gamma$ selector ({\em best opt. flow only (corrected})), and the overall accuracy on HMDB51 is quite close to ({\em all 4 opt. flow (corrected)}). Note that ({\em best opt. flow only (corrected)}) only hallucinate one best OFF, whereas ({\em all 4 opt. flow (corrected)}) hallucinate 4 best OFFs and each best optical flow feature is chosen from each optical flow. 

For the rest experiments, by default, we choose to hallucinate the top performing optical flow feature selected from all 4 OFFs by using our selector ({\em best opt. flow only (corrected})).

\vspace{-0.2cm}
\subsection{Comparisons With the SOTA Methods}
\label{sec:comparisons}


\begin{table}[t]
\begin{center}
\setlength{\tabcolsep}{0.12em}
\renewcommand{\arraystretch}{0.70}
\fontsize{9}{9}\selectfont
\resizebox{\linewidth}{!}{\begin{tabular}{ l c c c c c c}
\toprule
& IDT-FV~\cite{improved_traj} & Two-stream net.~\cite{christoph2016cvpr} & I3D~\cite{i3d_net} & DEEP-HAL~\cite{Wang_2019_ICCV} & ODF+SDF~\cite{leimm21} & HAL (ours)\\
\hline
Original & 57.2 & 69.2 & 80.9 & 82.5 & 87.6 & 83.5 \\
\rowcolor{LightCyan}
{\em Flow Corr.} (ours)  & 60.7  & 77.8  & 83.0  & 85.0  & {\bf 89.0}  & 87.3 \\
Improvement & \textcolor{red}{$\uparrow$\textbf{3.5}} & \textcolor{red}{$\uparrow$\textbf{8.6}} & \textcolor{red}{$\uparrow$\textbf{2.1}} & \textcolor{red}{$\uparrow$\textbf{2.5}} & \textcolor{red}{$\uparrow$\textbf{1.4}} & \textcolor{red}{$\uparrow$\textbf{3.8}} \\
\end{tabular}}
%
\setlength{\tabcolsep}{0.12em}
\renewcommand{\arraystretch}{0.70}
\fontsize{9}{9}\selectfont
\resizebox{\linewidth}{!}{\begin{tabular}{ c c c c c}
\midrule
\kern-0.5em EvaNet \cite{Piergiovanni_2019_ICCV} $82.3$ & BIKE~\cite{bike} $84.3$ & SCK+DCK~\cite{piotr2019} $86.1$ & VideoMAE V2~\cite{wang2023videomaev2} $88.1$ \\
\bottomrule
\end{tabular}}
%
\vspace{-0.5cm}
\caption{Evaluations of various methods ({\em top}) w/wo our flow dynamics correction and ({\em bottom}) comparisons
to the state of the art on  HMDB-51. 
}
\label{tab:hmdb51f}
\end{center}
\vspace{-0.5cm}
\end{table}


Tab.~\ref{tab:hmdb51f} shows the results on HMDB-51. With corrected flow dynamics (denoted as {\em Flow Corr.}), IDT-FV outperforms the use of original optical flow by 3.5\%. The use of optical flow correction on two-stream network boosts the performance by more than $8\%$. Although our HAL uses only the optical flow streams, it still achieves very competitive results compared to its similar competitors, \eg, DEEP-HAL and ODF+SDF.


\begin{table}[t]
\begin{center}
\setlength{\tabcolsep}{0.12em}
\renewcommand{\arraystretch}{0.70}
\fontsize{9}{9}\selectfont
\resizebox{\linewidth}{!}{\begin{tabular}{ l c c c c c}
\toprule
& Two-stream net.~\cite{christoph2016cvpr} & I3D~\cite{i3d_net} & ADL I3D~\cite{anoop_advers} & DEEP-HAL~\cite{Wang_2019_ICCV} & HAL (ours)\\
\hline
Original & 92.0 / 91.9 & 89.9 / - & 91.7 / - & 92.2 / 92.6 & 92.4 / 92.6 \\
\rowcolor{LightCyan}
{\em Flow Corr.} (ours) & 92.4 / 92.8 & 90.3 / -  & 92.2 / -  & 92.4 / 92.8  & {\bf 94.3} / {\bf 94.2} \\
Improvement & \textcolor{red}{$\uparrow$\textbf{0.9}} & \textcolor{red}{$\uparrow$\textbf{0.4}} & \textcolor{red}{$\uparrow$\textbf{0.5}} & \textcolor{red}{$\uparrow$\textbf{0.2}}& \textcolor{red}{$\uparrow$\textbf{1.9}} \\
\midrule
T-ResNet \cite{yuppp} & $87.0$ / $87.6$ & & MSOE(two-stream) \cite{Hadji_2018_ECCV} & $92.0$ / $91.9$ & \\
\bottomrule
\end{tabular}}
\vspace{-0.3cm}
\caption{Evaluations of various methods ({\em top}) w/wo our flow dynamics correction and ({\em bottom}) comparisons to the state of the art on YUP++. We report {\em mean over stat.\&dyn.} / {\em mean over all (stat.\&dyn.\&mixed)}. 
}
\label{tab:yupf}
\end{center}
\vspace{-0.5cm}
\end{table}



Tab.~\ref{tab:yupf} shows the results on YUP++. We notice that our HAL performs equally well compared to DEEP-HAL even without the use of flow dynamics correction, and with our corrected optical flow, it outperforms the baseline method (DEEP-HAL) by $\sim$2\%. Our method also outperforms more complex T-ResNet and MSOE (two-stream) by $>2\%$.


\begin{table}[!tb]
\begin{center}
\setlength{\tabcolsep}{0.12em}
\renewcommand{\arraystretch}{0.70}
\resizebox{0.85\linewidth}{!}{\begin{tabular}{ l c c c c c}
\toprule
& IDT-FV~\cite{improved_traj} & I3D~\cite{i3d_net} & DEEP-HAL~\cite{Wang_2019_ICCV} & ODF+SDF~\cite{leimm21} & HAL (ours)\\
\hline
Original & 67.6 & 74.8 & 81.8 & 84.8 & 82.8 \\
\rowcolor{LightCyan}
{\em Flow Corr.} (ours) & 74.0 & 80.4  & 83.5  & {\bf 86.2}  & {\bf 86.2} \\
Improvement & \textcolor{red}{$\uparrow$\textbf{6.4}} & \textcolor{red}{$\uparrow$\textbf{5.6}} & \textcolor{red}{$\uparrow$\textbf{1.7}} &\textcolor{red}{$\uparrow$\textbf{1.4}} & \textcolor{red}{$\uparrow$\textbf{3.4}}\\
\end{tabular}}
%
\setlength{\tabcolsep}{0.12em}
\renewcommand{\arraystretch}{0.70}
\fontsize{9}{9}\selectfont
\resizebox{0.85\linewidth}{!}{\begin{tabular}{ c c c c}
\midrule
\kern-0.5em KRP-FS \cite{anoop_rankpool_nonlin} $70.0$  & KRP-FS+IDT \cite{anoop_rankpool_nonlin} $76.1$  & GRP \cite{anoop_generalized} $68.4$  & GRP+IDT \cite{anoop_generalized} $75.5$ \kern-0.5em\\
\bottomrule
\end{tabular}}
\vspace{-0.3cm}
\caption{Evaluations of various methods ({\em top}) w/wo our flow dynamics correction and ({\em bottom}) comparisons to the state of the art on MPII. 
}
\label{tab:mpiif}
\end{center}
\vspace{-0.5cm}
\end{table}

Tab.~\ref{tab:mpiif} shows that our HAL achieves on par mAP performance compared to more complicated ODF+SDF when we activate the use of flow correction in its optical flow stream, and it performs better than DEEP-HAL by $\sim$ 3\%. 
Flow dynamics correction boosts IDT-FV, I3D, DEEP-HAL and ODF+SDF for AR by $\sim$ 6\%, 6\%, 2\% and 2\% respectively on MPII. 
Note that MPII contains some fine-grained actions where different motion dynamics are of great importance to the recognition tasks, and our model with optical flow correction achieves the new state-of-the-art performance.

\begin{table}[!tb]
\begin{center}
\setlength{\tabcolsep}{0.12em}
\renewcommand{\arraystretch}{0.70}
\resizebox{\linewidth}{!}{\begin{tabular}{ l c c c c c c}
\toprule
& I3D~\cite{i3d_net} & DEEP-HAL~\cite{Wang_2019_ICCV} & AssembleNet~\cite{assemblenet} & AssembleNet++~\cite{assemblenet_plus} & HAL (ours,  & HAL (ours, \\
& & & & & (with I3D) & AssembleNet++)\\
\hline
Original & 40.0 & 43.1 & 56.6 & 59.8 & 45.3 & {\bf 62.0} \\
\rowcolor{LightCyan}
{\em Flow Corr.} (ours) & 42.1  & 45.7  & 59.7  & 62.0  & 48.7  & {\bf 64.9} \\
Improvement & \textcolor{red}{$\uparrow$\textbf{2.1}} & \textcolor{red}{$\uparrow$\textbf{2.6}} & \textcolor{red}{$\uparrow$\textbf{3.1}}& \textcolor{red}{$\uparrow$\textbf{2.2}} & \textcolor{red}{$\uparrow$\textbf{3.4}} & \textcolor{red}{$\uparrow$\textbf{2.9}}\\
\end{tabular}}
%
\setlength{\tabcolsep}{0.12em}
\renewcommand{\arraystretch}{0.70}
\fontsize{9}{9}\selectfont
\resizebox{\linewidth}{!}{\begin{tabular}{ c c c c c}
\midrule
\kern-0.5em ActionCLIP~\cite{DBLP:journals/corr/abs-2109-08472} $44.3$ & SlowFast~\cite{slowfast} $45.2$ & En-VidTr-L~\cite{Zhang_2021_ICCV} $47.3$  & MoViNet-A6~\cite{kondratyuk2021movinets} $63.2$ &  TubeViT-L~\cite{piergiovanni2023rethinking} $66.2$ \\
\bottomrule
\end{tabular}
}
\vspace{-0.5cm}
\caption{Evaluations of various methods ({\em top}) w/wo flow dynamics correction and ({\em bottom}) comparisons to the state of the art on Charades. 
}
\label{tab:charadesff}
\end{center}
\vspace{-0.5cm}
\end{table}

Tab.~\ref{tab:charadesff} shows our simple HAL achieves the best results on Charades.
Our model is based on self-supervision, which learns to hallucinate the best motion dynamic features, makes our pipeline lightweight in comparison to competitors such as Contrastive Language-Image Pre-Training (CLIP) model, e.g., ActionCLIP, and video transformer-based model, e.g., En-VidTr-L. 
With AssembleNet++ backbone and our flow dynamics correction, we outperform AssembleNet++ by $\sim$ 3\%.


\section{Conclusions}
\label{sec:concl}
In this paper, we address the challenge of selecting and enhancing motion dynamics through power normalizing optical flow speed components, achieving state-of-the-art results in action recognition benchmarks. Our approach allows tailored modeling of actions based on their significance (e.g., distinguishing subtle hand waves from robust walks or jogs). We show that leading action recognition methods benefit from our flow dynamics correction, and our low-computational-cost pipeline is advantageous for tasks like clustering and captioning in video processing.

{\small
\bibliographystyle{ieee}
\bibliography{egbib}
}


\end{document}